\documentclass{llncs}
\usepackage{epic,eepic,epsfig}
\usepackage{graphicx}
\usepackage{amsfonts}
\usepackage{times}
\usepackage{helvet}
\usepackage{courier}
\newcommand{\Sol}{\textit{\rmfamily Sol}}
\newcommand{\CalculateValidDomains}{\textit{\rmfamily CVD}}
\newcommand{\dom}{\textrm{\rmfamily dom}}

\title{Calculating Valid Domains for BDD-Based Interactive Configuration}

\author{Tarik Hadzic, Rune Moller Jensen, Henrik Reif Andersen}
\institute{Computational Logic and Algorithms Group, \\
IT University of Copenhagen, Denmark \\
\email{tarik@itu.dk,rmj@itu.dk,hra@itu.dk}}

\begin{document}

%\makefront
 \maketitle

\begin{abstract}
In these notes we formally describe the functionality of
Calculating Valid Domains from the BDD representing the solution
space of valid configurations. The formalization is largely based
on the CLab \cite{CLAB} configuration framework.

\end{abstract}

\section{Introduction}

Interactive configuration problems are special applications of
Constraint Satisfaction Problems (CSP) where a user is assisted in
interactively assigning values to variables by a software  tool.
This software, called a configurator, assists the user by
calculating and displaying the available, valid choices for each
unassigned variable in what are called \emph{valid domains
computations}. Application areas include customising physical
products (such as PC's and cars) and services (such as airplane
tickets and insurances).

Three important features are required of a tool that implements
interactive configuration: it should be complete (all valid
configurations should be reachable through user interaction),
backtrack-free (a user is never forced to change an earlier choice
due to incompleteness in the logical deductions), and it should
provide real-time performance (feedback should be fast enough to
allow real-time interactions).  The requirement of obtaining
backtrack-freeness while maintaining completeness makes the
problem of calculating valid domains NP-hard. The real-time
performance requirement enforces further that runtime calculations
are bounded in polynomial time.   According to user-interface
design criteria, for a user to perceive interaction  as being
real-time, system response needs to be within about 250
milliseconds in practice \cite{Rask:00:HumanInterface}. Therefore,
the current approaches that meet all three conditions use off-line
precomputation to generate an efficient runtime data structure
representing the solution space \cite{AFM02,M03,HSJAMH04,MAH01}.
The challenge with this data structure is that the solution space
is almost always exponentially large and it is NP-hard to find.
Despite the bad worst-case bounds, it has nevertheless turned out
in real industrial applications that the data structures can often
be kept small \cite{CS04,HSJAMH04,M03}.

\section{Interactive Configuration} \label{sec:InterConf}
 The input \emph{model} to an interactive configuration problem is
 a special kind of Constraint Satisfaction Problem (CSP)
\cite{Tsang93,Dech03} where constraints are represented as
propositional formulas:

\begin{definition}
A \emph{configuration model} $C$ is a triple $(X,D,F)$ where X is
a set of variables $\{x_0, \ldots, x_{n-1}\}$, $D = D_0 \times
\ldots \times D_{n-1}$ is the Cartesian product of their finite
domains $D_0, \ldots ,D_{n-1}$ and $F = \{f_0, . . . , f_{m-1}\}$
is a set of propositional formulae over atomic propositions $x_i =
v$, where $v \in D_i$, specifying conditions on the values of the
variables.
\end{definition}

Concretely, every domain can be defined as $D_i = \{0, \ldots,
|D_i|-1 \}$. An assignment of values $v_0, \ldots, v_{n-1}$ to
variables $x_0, \ldots, x_{n-1}$ is denoted as an assignment $\rho
= \{(x_0,v_0),\ldots,(x_{n-1},v_{n-1})\}$. Domain of assignment
$dom(\rho)$ is the set of variables which are assigned: $dom(\rho)
= \{x_i \mid \exists v\in D_i.(x_i,v) \in \rho \}$ and if
$dom(\rho)=X$ we refer to $\rho$ as a  \emph{total assignment}.
%hence for total assignment $dom(\rho)=X$.
We say that a total assignment $\rho$ is \emph{valid}, if it
satisfies all the rules which is denoted as $\rho \models F$.
% represents a fully

A partial assignment $\rho', dom(\rho') \subseteq X$ is
\emph{valid} if there is at least one total assignment $\rho
\supseteq \rho'$ that is valid $\rho \models F$, i.e. if there is
at least one way to successfully finish the existing configuration
process.

%In product configuration, the knowledge about product components
%and product rules is usually modelled by representing all the
%choices for a component as values in a variable domain. Then, a
%valid assignment $\rho$ completely specifies a configurable
%product.

%, where for $i$-th component type a specific
%%component $v_i$ has been chosen.
%The configured product must satisfy all the product rules, which
%are captured by propositional formulae $F$. Hence, the user
%specified assignment
%
%If only a subset of components has been specified, then we have a
\begin{example}
Consider specifying a T-shirt by choosing the color (black, white,
red, or blue), the size (small, medium, or large) and the print
("Men In Black" - MIB or "Save The Whales" - STW). There are two
rules that we have to observe: if we choose the MIB print then the
color black has to be chosen as well, and if we choose the small
size then the STW print (including a big picture of a whale)
cannot be selected as the large whale does not fit on the small
shirt. The configuration problem $(X,D,F)$ of the T-shirt example
consists of variables $X = \{x_1,x_2,x_3\}$ representing color,
size and print. Variable domains are $D_1 = \{{\it black},{\it
white},{\it red},{\it blue}\}$, $D_2 = \{{\it small},{\it
medium},{\it large}\}$, and $D_3 = \{{\it MIB},{\it STW}\}$. The
two rules translate to $F = \{f_1,f_2\}$, where $f_1 = (x_3 = {\it
MIB}) \Rightarrow (x_1 = {\it black})$ and $f_2 = (x_3 = {\it
STW}) \Rightarrow (x_2 \neq {\it small})$. There are
$|D_1||D_2||D_3| = 24$ possible assignments. Eleven of these
assignments are valid configurations and they form the solution
space shown in Fig.~\ref{fig:solspace}.\hfill$\Diamond$
\end{example}

\begin{figure}
\begin{center}
\begin{tabular}{lllll}
$({\it black},{\it small},{\it MIB })$ & \hspace{5mm} & $({\it black},{\it large},{\it STW})$  & \hspace{5mm} & $({\it red},{\it large},{\it STW})$ \\
$({\it black},{\it medium},{\it MIB})$ & \hspace{5mm} & $({\it white},{\it medium},{\it STW})$ & \hspace{5mm} & $({\it blue},{\it medium},{\it STW})$ \\
$({\it black},{\it medium},{\it STW})$ & \hspace{5mm} & $({\it white},{\it large},{\it STW})$  & \hspace{5mm} & $({\it blue},{\it large},{\it STW})$ \\
$({\it black},{\it large},{\it MIB})$  & \hspace{5mm} & $({\it red},{\it medium},{\it STW})$   & \\
\end{tabular}
\caption{\label{fig:solspace} Solution space for the T-shirt
example}
\end{center}

\end{figure}

\subsection{User Interaction}
%Given a configuration model $C=(X,D,F)$ and a partial user
%assignment $\rho$, \emph{interactive configuration} is the process
%of assisting a user interactively to reach a total valid
%assignment, starting from $\rho$.
 Configurator assists a user interactively to reach a valid product specification, i.e. to
 reach total valid assignment. The key operation in this interaction is that of
computing, for each unassigned variable $x_i \in X \setminus
dom(\rho)$, the \emph{valid domain} $D_i^{\rho} \subseteq D_i$.
The domain is \emph{valid} if it contains those and only those
values with which $\rho$ can be extended to become a total valid
assignment, i.e. $D_i^{\rho} = \{v \in D_i \mid \exists \rho':
\rho' \models F \wedge \rho \cup \{(x_i,v)\} \subseteq \rho'\}$.
The significance of this demand is that it guarantees the user
backtrack-free assignment to variables as long as he selects
values from valid domains. This reduces cognitive effort during
the interaction and
increases usability.% For ease of presentation, we shall take
%$D_i^\rho=\{v_i\}$ for $(x_i,v_i)\in\rho$ and refer to the tuple
%of all the $D_i^\rho$'s as $D^\rho$.

At each step of the interaction, the configurator reports the
valid domains to the user, based on the current partial assignment
$\rho$ resulting from his earlier choices. The user then picks an
unassigned variable $x_j \in X \setminus dom(\rho)$ and selects a
value from the calculated valid domain $v_j \in D_j^{\rho}$. The
partial assignment is then extended to $\rho \cup \{(x_j,v_j)\}$
and a new interaction step is initiated.

\section{BDD Based Configuration} \label{sec:BDDConf} In \cite{HSJAMH04,SJHAHM} the
interactive configuration was delivered by dividing the
computational effort into an \emph{offline} and \emph{online}
phase. First, in the offline phase, the authors compiled a BDD
representing the solution space of all valid configurations $Sol =
\{\rho \mid \rho \models F\}$. Then, the functionality of
\emph{calculating valid domains} ($CVD$) was delivered online, by
efficient algorithms executing during the interaction with a user.
The benefit of this approach is that the BDD needs to be compiled
only once, and can be reused for multiple user sessions. The user
interaction process is illustrated in Fig. \ref{fig:InCo}.

\begin{figure}[htbp]
\ttfamily

\ \ \    $InCo(\Sol,\rho)$

\ \ \   1:  \ \ \ \   while $|\Sol^\rho|>1$

\ \ \   2:  \ \ \ \   \ \ \ \    compute
$D^\rho=\CalculateValidDomains(\Sol,\rho)$

\ \ \   3:  \ \ \ \   \ \ \ \   report $D^\rho$ to the user

\ \ \   4:  \ \ \ \   \ \ \ \   the user chooses $(x_i, v)$ for
some $x_i\not\in \dom(\rho)$, $v \in D_i^\rho$

\ \ \   5:  \ \ \ \   \ \ \ \   $\rho \leftarrow \rho\cup{\{(x_i,
v)\}}$

\ \ \   6:  \ \ \ \   return   $\rho$

 \rmfamily
 \caption{\label{fig:InCo} Interactive configuration algorithm working on a
 BDD representation of the solutions $\Sol$ reaches a valid total configuration
 as an extension of the argument $\rho$.}
\end{figure}

Important requirement for online user-interaction is the
guaranteed real-time experience of user-configurator interaction.
Therefore, the algorithms that are executing in the online phase
must be provably efficient in the size of the BDD representation.
This is what we call the \emph{real-time guarantee}. As the $CVD$
functionality is NP-hard, and the online algorithms are polynomial
in the size of generated BDD, there is no hope of providing
polynomial  size guarantees for  the worst-case  BDD
representation. However, it suffices that the BDD size is small
enough for all the configuration instances occurring in practice
\cite{SJHAHM}.

% obtained by compiling the set of valid configurations
%$Sol = \{\rho \mid \rho \models F \}$ into a Binary Decision
%Diagram (BDD) \cite{B86} during the offline phase. Then the $CVD$
%functionality was efficiently executed on top of this
%representation during user interaction.

%In this section we formally describe the BDD-based $CVD$
%algorithms, largely based on the Clab \cite{CLAB} configuration
%framework. %This formalization will be a basis for describing our
%%extended algorithms that allow maximum price manipulation.

\subsection{Binary Decision Diagrams}
A reduced ordered Binary Decision Diagram (BDD) is a rooted
directed acyclic graph representing a Boolean function on a set of
linearly ordered Boolean variables. It has one or two terminal
nodes labeled 1 or 0 and a set of variable nodes. Each variable
node is associated with a Boolean variable and has two outgoing
edges {\em low} and {\em high}. Given an assignment of the
variables, the value of the Boolean function is determined by a
path starting at the root node and recursively following the high
edge, if the associated variable is true, and the low edge, if the
associated variable is false. The function value is {\em true}, if
the label of the reached terminal node is 1; otherwise it is {\em
false}. The graph is ordered such that all paths respect the
ordering of the variables.

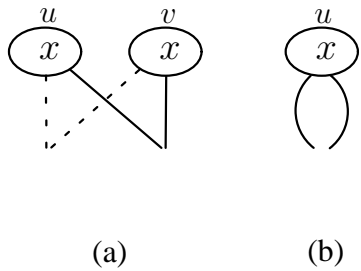
\begin{figure}
 \centering \setlength{\unitlength}{0.00083333in}
\begingroup\makeatletter\ifx\SetFigFont\undefined%
\gdef\SetFigFont#1#2#3#4#5{%
  \reset@font\fontsize{#1}{#2pt}%
  \fontfamily{#3}\fontseries{#4}\fontshape{#5}%
  \selectfont}%
\fi\endgroup%
{\renewcommand{\dashlinestretch}{30}
\begin{picture}(2205,1704)(0,-10)
\thicklines
\put(2098.990,1003.170){\arc{594.038}{2.2388}{3.9599}}
\put(1838.869,995.785){\arc{580.264}{5.3999}{7.1973}}
\put(1965,1370){\ellipse{450}{300}}
\put(240,1370){\ellipse{450}{300}}
\put(990,1370){\ellipse{450}{300}}
\path(994,1216)(990,770)
\dashline{45.000}(245,1220)(247,770)
\path(394,1258)(964,770)
\dashline{45.000}(828,1256)(277,770)
\put(205,1557){\makebox(0,0)[lb]{\smash{{{\SetFigFont{12}{14.4}{\rmdefault}{\mddefault}{\itdefault}$u$}}}}}
\put(959,1549){\makebox(0,0)[lb]{\smash{{{\SetFigFont{12}{14.4}{\rmdefault}{\mddefault}{\itdefault}$v$}}}}}
\put(1919,1546){\makebox(0,0)[lb]{\smash{{{\SetFigFont{12}{14.4}{\rmdefault}{\mddefault}{\itdefault}$u$}}}}}
\put(202,1325){\makebox(0,0)[lb]{\smash{{{\SetFigFont{14}{16.8}{\rmdefault}{\mddefault}{\itdefault}$x$}}}}}
\put(952,1325){\makebox(0,0)[lb]{\smash{{{\SetFigFont{14}{16.8}{\rmdefault}{\mddefault}{\itdefault}$x$}}}}}
\put(1927,1325){\makebox(0,0)[lb]{\smash{{{\SetFigFont{14}{16.8}{\rmdefault}{\mddefault}{\itdefault}$x$}}}}}
\put(533,54){\makebox(0,0)[lb]{\smash{{{\SetFigFont{12}{14.4}{\rmdefault}{\mddefault}{\updefault}(a)}}}}}
\put(1875,62){\makebox(0,0)[lb]{\smash{{{\SetFigFont{12}{14.4}{\rmdefault}{\mddefault}{\updefault}(b)}}}}}
\end{picture}
}
 \begin{center}
   \caption{\label{fig:bddEx}(a) nodes associated to the same variable
            with equal low and high successors will be converted to a
            single node. (b) nodes causing redundant tests on a variable
            are eliminated. High and low edges are drawn with solid and
            dashed lines, respectively}
 \end{center}
\end{figure}

A BDD is reduced such that no pair of distinct nodes $u$ and $v$
are associated with the same variable and low and high successors
(Fig.~\ref{fig:bddEx}a), and no variable node $u$ has identical
low and high successors (Fig.~\ref{fig:bddEx}b).
Due to these reductions, the number of nodes in a BDD for many
functions encountered in practice is often much smaller than the
number of truth assignments of the function. Another advantage is
that the reductions make BDDs canonical \cite{B86}. Large space
savings can be obtained by representing a collection of BDDs in a
single multi-rooted graph where the sub-graphs of the BDDs are
shared. Due to the canonicity, two BDDs are identical if and only
if they have the same root. Consequently, when using this
representation, equivalence checking between two BDDs can be done
in constant time. In addition, BDDs are easy to manipulate. Any
Boolean operation on two BDDs can be carried out in time
proportional to the product of their size. The size of a BDD can
depend critically on the variable ordering. To find an optimal
ordering is a co-NP-complete problem in itself \cite{B86}, but a
good heuristic for choosing an ordering is to locate dependent
variables close to each other in the ordering. For a comprehensive
introduction to BDDs and {\em branching programs} in general, we
refer the reader to Bryant's original paper \cite{B86} and the
books \cite{MT98,W00}.

\subsection{Compiling the Configuration Model}
 Each of the finite domain variables $x_i$ with domain
$D_i = \{0, \ldots, |D_i|-1 \}$ is encoded by $k_i = \lceil
log|D_i| \rceil$ Boolean variables $x_0^i, \ldots, x_{k_i-1}^i$.
Each $j \in D_i$, corresponds to a binary encoding  $\overline{v_0
\ldots v_{k_i-1}}$ denoted as $v_0 \ldots v_{k_i-1} = enc(j)$.
Also, every combination of Boolean values $v_0 \ldots v_{k_i-1}$
represents some integer $j \leq 2^{k_i}-1$, denoted as $j =
dec(v_0 \ldots v_{k_i-1})$. Hence, atomic proposition $x_i = v$ is
encoded as a Boolean expression $x_0^i = v_0 \wedge \ldots \wedge
x_{k_i-1}^i = v_{k_i-1}$. In addition, \emph{domain constraints}
are added to forbid those assignments to $v_0 \ldots v_{k_i-1}$
which do not translate to a value in $D_i$, i.e. where $dec(v_0
\ldots v_{k_i-1}) \geq |D_i|$.

%\subsubsection{1-Hot encoding} For every value $j \in D_i$ we
%introduce a Boolean variable $x_j^i$. Now, $x_i = j$ is encoded as
%$x_j^i = 1$. Since only one of the values for a variable $x_i$ can
%be selected, we have $x_j^i=1 \Rightarrow \bigwedge_{l \in D_i
%\setminus \{j\}}x_l^i=0$, $0 \leq j \leq |D_i|-1$.

% ---------------------- FORMALIZATION, ALGORITHMS -----------------
Let the solution space $Sol$ over ordered set of variables $x_0 <
\ldots < x_{k-1}$
 be represented by a Binary Decision Diagram $B(V,E,X_b,R,var)$,
where $V$ is the set of nodes $u$, $E$ is the set of edges $e$ and
$X_b = \{0, 1, \ldots, |X_b|-1\}$ is an ordered set of variable
indexes, labelling every non-terminal node $u$ with $var(u) \leq
|X_b|-1$ and labelling the terminal nodes $T_0,T_1$ with index
$|X_b|$. Set of variable indexes $X_b$ is constructed by taking
the union of  Boolean encoding variables
$\bigcup_{i=0}^{n-1}\{x_0^i, \ldots, x_{k_i-1}^i\}$ and ordering
them in a natural layered way, i.e. $x_{j_1}^{i_1} <
x_{j_2}^{i_2}$ iff $i_1 < i_2$ or $i_1=i_2$ and $j_1<j_2$.

Every directed  edge $e = (u_1, u_2)$ has a starting vertex $u_1 =
\pi_1(e)$ and ending vertex $u_2 = \pi_2(e)$. $R$ denotes the root
node of the BDD.

%Every edge $e \in E$ where $e.u_2 = high(e.v_1)$ represents an
%assignment $x_j = 1$ where $j = var(e.v_1)$ hence making  a
%component $j$ part of the final product. Hence, $1 \in D_j$.

\begin{figure}
 \centering \setlength{\unitlength}{0.00083333in}
\begingroup\makeatletter\ifx\SetFigFont\undefined%
\gdef\SetFigFont#1#2#3#4#5{%
  \reset@font\fontsize{#1}{#2pt}%
  \fontfamily{#3}\fontseries{#4}\fontshape{#5}%
  \selectfont}%
\fi\endgroup%
{\renewcommand{\dashlinestretch}{30}
\begin{picture}(2642,3313)(0,-10)
\thicklines
\path(1466,2984)(1564,2099)
\blacken\path(1535.033,2156.159)(1564.000,2099.000)(1579.760,2161.112)(1535.033,2156.159)
\put(739,1934){\ellipse{450}{300}}
\put(964,1334){\ellipse{450}{300}}
\put(1564,1334){\ellipse{450}{300}}
\put(364,1334){\ellipse{450}{300}}
\put(1939,734){\ellipse{450}{300}}
\put(1039,2534){\ellipse{450}{300}}
\put(1414,3134){\ellipse{450}{300}}
\put(364,734){\ellipse{450}{300}}
\put(2164,1334){\ellipse{450}{300}}
\put(1564,1934){\ellipse{450}{300}}
\dashline{90.000}(1346,2984)(1174,2669)
\blacken\path(1183.007,2732.444)(1174.000,2669.000)(1222.502,2710.878)(1183.007,2732.444)
\path(1129,2377)(1421,2077)
\blacken\path(1363.027,2104.302)(1421.000,2077.000)(1395.274,2135.689)(1363.027,2104.302)
\dashline{90.000}(979,2354)(829,2092)
\blacken\path(839.285,2155.249)(829.000,2092.000)(878.337,2132.891)(839.285,2155.249)
\dashline{90.000}(664,1777)(476,1484)
\blacken\path(489.465,1546.649)(476.000,1484.000)(527.339,1522.348)(489.465,1546.649)
\dashline{90.000}(356,1184)(356,899)
\blacken\path(333.500,959.000)(356.000,899.000)(378.500,959.000)(333.500,959.000)
\path(796,1768)(956,1484)
\blacken\path(906.946,1525.231)(956.000,1484.000)(946.152,1547.319)(906.946,1525.231)
\dashline{90.000}(1549,1777)(1549,1507)
\blacken\path(1526.500,1567.000)(1549.000,1507.000)(1571.500,1567.000)(1526.500,1567.000)
\dashline{90.000}(334,577)(724,299)
\blacken\path(662.082,315.505)(724.000,299.000)(688.202,352.149)(662.082,315.505)
\path(581,689)(1354,299)
\blacken\path(1290.297,305.939)(1354.000,299.000)(1310.567,346.115)(1290.297,305.939)
\path(1045,1173)(1466,299)
\blacken\path(1419.691,343.291)(1466.000,299.000)(1460.233,362.820)(1419.691,343.291)
\dashline{90.000}(924,1186)(926,307)
\blacken\path(903.364,366.949)(926.000,307.000)(948.363,367.051)(903.364,366.949)
\dashline{90.000}(1549,1177)(1549,307)
\blacken\path(1526.500,367.000)(1549.000,307.000)(1571.500,367.000)(1526.500,367.000)
\path(1646,1177)(1856,892)
\blacken\path(1802.294,926.956)(1856.000,892.000)(1838.522,953.650)(1802.294,926.956)
\dashline{90.000}(2074,1177)(1946,884)
\blacken\path(1949.401,947.990)(1946.000,884.000)(1990.638,929.975)(1949.401,947.990)
\dashline{90.000}(1924,569)(1639,307)
\blacken\path(1667.944,364.171)(1639.000,307.000)(1698.399,331.042)(1667.944,364.171)
\path(1796,607)(994,314)
\blacken\path(1042.636,355.723)(994.000,314.000)(1058.078,313.455)(1042.636,355.723)
\path(1708,1809)(2051,1484)
\blacken\path(1991.971,1508.936)(2051.000,1484.000)(2022.922,1541.601)(1991.971,1508.936)
\put(2089,1273){\makebox(0,0)[lb]{{\SetFigFont{11}{13.2}{\rmdefault}{\mddefault}{\updefault}$x_2^1$}}}
\put(1872,673){\makebox(0,0)[lb]{{\SetFigFont{11}{13.2}{\rmdefault}{\mddefault}{\updefault}$x_3^0$}}}
\put(672,1881){\makebox(0,0)[lb]{{\SetFigFont{11}{13.2}{\rmdefault}{\mddefault}{\updefault}$x_2^0$}}}
\put(1497,1867){\makebox(0,0)[lb]{{\SetFigFont{11}{13.2}{\rmdefault}{\mddefault}{\updefault}$x_2^0$}}}
\put(290,1273){\makebox(0,0)[lb]{{\SetFigFont{11}{13.2}{\rmdefault}{\mddefault}{\updefault}$x_2^1$}}}
\put(889,1266){\makebox(0,0)[lb]{{\SetFigFont{11}{13.2}{\rmdefault}{\mddefault}{\updefault}$x_2^1$}}}
\put(1482,1266){\makebox(0,0)[lb]{{\SetFigFont{11}{13.2}{\rmdefault}{\mddefault}{\updefault}$x_2^1$}}}
\put(282,666){\makebox(0,0)[lb]{{\SetFigFont{11}{13.2}{\rmdefault}{\mddefault}{\updefault}$x_3^0$}}}
\put(1331,3082){\makebox(0,0)[lb]{{\SetFigFont{11}{13.2}{\rmdefault}{\mddefault}{\updefault}$x_1^0$}}}
\put(965,2474){\makebox(0,0)[lb]{{\SetFigFont{11}{13.2}{\rmdefault}{\mddefault}{\updefault}$x_1^1$}}}
\path(259,1199)(258,1198)(256,1195)
	(251,1191)(245,1184)(236,1174)
	(225,1162)(212,1147)(197,1129)
	(180,1110)(162,1088)(144,1064)
	(126,1039)(108,1013)(91,987)
	(75,959)(61,930)(48,901)
	(38,870)(30,839)(24,806)
	(22,772)(24,736)(30,698)
	(40,660)(56,622)(75,587)
	(97,554)(123,523)(150,494)
	(179,467)(209,442)(241,420)
	(273,398)(307,379)(341,360)
	(375,343)(410,327)(445,311)
	(480,297)(514,283)(548,271)
	(579,259)(609,249)(636,239)
	(660,231)(681,225)(698,219)
	(711,215)(731,209)
\blacken\path(667.065,204.690)(731.000,209.000)(679.996,247.792)(667.065,204.690)
\path(2201,1177)(2202,1175)(2203,1172)
	(2206,1166)(2210,1156)(2215,1143)
	(2222,1126)(2230,1106)(2238,1083)
	(2247,1057)(2257,1029)(2265,1000)
	(2274,969)(2281,937)(2287,905)
	(2292,871)(2295,837)(2295,802)
	(2293,766)(2288,728)(2280,689)
	(2268,649)(2252,609)(2231,569)
	(2206,531)(2179,495)(2149,463)
	(2118,435)(2087,409)(2055,386)
	(2022,366)(1989,348)(1955,331)
	(1922,317)(1888,303)(1855,291)
	(1823,280)(1792,270)(1763,261)
	(1736,253)(1713,247)(1694,242)
	(1678,238)(1654,232)
\blacken\path(1706.751,268.380)(1654.000,232.000)(1717.666,224.724)(1706.751,268.380)
\path(1368,285)(1646,285)(1646,22)
	(1368,22)(1368,285)
\path(746,285)(1024,285)(1024,22)
	(746,22)(746,285)
\put(844,89){\makebox(0,0)[lb]{{\SetFigFont{14}{16.8}{\rmdefault}{\mddefault}{\updefault}1}}}
\put(1459,89){\makebox(0,0)[lb]{{\SetFigFont{14}{16.8}{\rmdefault}{\mddefault}{\updefault}0}}}
\end{picture}
}
 \begin{center}
   \caption{\label{fig:shirtBDD}BDD of the solution space of the T-shirt example. Variable $x_i^j$ denotes bit $v_j$
   of the Boolean encoding of finite domain variable $x_i$.}
 \end{center}
\end{figure}
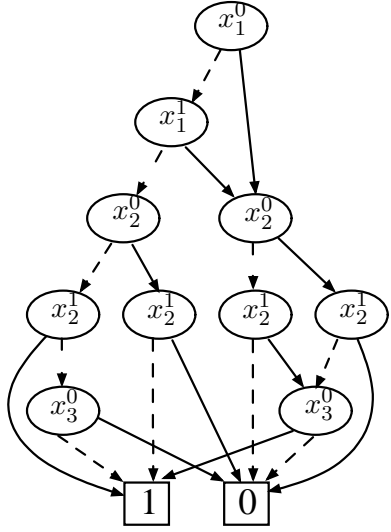

\begin{example}
The BDD representing the solution space of the T-shirt example
introduced in Sect.~\ref{sec:InterConf} is shown in
Fig.~\ref{fig:shirtBDD}. In the T-shirt example there are three
variables: $x_1, x_2$ and $x_3$, whose domain sizes are four,
three and two, respectively. Each variable is represented by a
vector of Boolean variables. In the figure the Boolean vector for
the variable $x_i$ with domain $D_i$ is $(x_i^{0},x_i^{1},\cdots
x_{i}^{l_i-1})$, where $l_i = \lceil \lg|D_i| \rceil$. For
example, in the figure, variable $x_2$ which corresponds to the
size of the T-shirt is represented by the Boolean vector
$(x_2^{0},x_2^{1})$. In the BDD any path from the root node to the
terminal node $1$, corresponds to one or more valid
configurations. For example, the path from the root node to the
terminal node $1$, with all the variables taking low values
represents the valid configuration $({\it black},{\it small},{\it
MIB })$. Another path with $x_1^0, x_1^1,$ and $ x_2^0$ taking low
values, and $x_2^1$ taking high value represents two valid
configurations: $({\it black},{\it medium},{\it MIB })$ and $({\it
black},{\it medium},{\it STW })$, namely. In this path the
variable $x_3^0$ is a don't care variable and hence can take both
low and high value, which leads to two valid configurations. Any
path from the root node to the terminal node $0$ corresponds to
invalid configurations. \hfill$\Diamond$
\end{example}

\section{Calculating Valid Domains}
Before showing the algorithms, let us first introduce the
appropriate notation. If an index $k \in X_b$ corresponds to the
$j+1$-st Boolean variable $x_j^i$ encoding the finite domain
variable $x_i$, we define $var_1(k) = i$ and $var_2(k) = j$ to be
the appropriate mappings.
% $X_b \rightarrow \{0, \ldots, n-1\}$ such that $var_1(k)
%= i$. Also, we define the function $var_2$ to take the value
%$var_2 \rightarrow $
%
%
%the mapping that associates every Boolean BDD variable $y_0,
%\ldots, y_{k-1}$, representing Boolean encoding variable $x_j^i$,
%to the index $i$ of corresponding finite domain variable $x_i$,
%and let $var_2$ map $x_j^i$ to the offset $j$.
Now, given the BDD $B(V,E,X_b,R,var)$, $V_i$ denotes the set of
all nodes $u \in V$ that are labelled with a BDD variable encoding
the finite domain variable $x_i$, i.e. $V_i = \{ u \in V \mid
var_1(u) = i\}$. We think of $V_i$ as defining a layer in the BDD.
We define $In_i$ to be the set of nodes $u \in V_i$ reachable by
an edge originating from outside the $V_i$ layer, i.e. $In_i = \{u
\in V_i | \ \exists (u',u) \in E. \ var_1(u')<i \}$. For the root
node $R$, labelled with $i_0 = var_1(R)$ we define $In_{i_0} =
V_{i_0}
= \{R\}$. %\emph{We assume that all the variables $X$ are relevant
%in $B$, i.e. every variable is labelling at least one node. In
%that case, $i_0 = var_1(R) = 0$.}

%Also define $Span_i$ to be the set of those nodes $u' \in
%V_j, j<i$ from which an edge $e$ originates such that spans across
%all the nodes $V_i$, i.e. $Span_i = \{u \in V_j | \ j<i \ and \
%\exists e=(v_1,v_2)\in E(B). \ \ u = v_1 \ and \ var_1(v_2) > i\}$

%We assume has fixed a variable value $$ s, denoted in partial
%assignment $\rho$.

We assume that in the previous user assignment, a user fixed a
value for a finite domain variable $x=v, x \in X$, extending the
old partial assignment $\rho_{old}$ to the current assignment
$\rho = \rho_{old} \cup \{(x,v)\}$. For every variable $x_i \in
X$, old valid domains are denoted as $D_i^{\rho_{old}},
i=0,\ldots,n-1$. and the old BDD $B^{\rho_{old}}$ is reduced to
the restricted BDD, $B^\rho(V,E,X_b,var)$. The $CVD$ functionality
is to calculate valid domains $D_i^\rho$ for remaining  unassigned
variables $x_i \not \in dom(\rho)$ by extracting values from the
newly restricted BDD $B^\rho(V,E,X_b,var)$.

To simplify the following discussion, we will analyze the isolated
execution of the $CVD$ algorithms over a given BDD
$B(V,E,X_b,var)$. The task is to calculate valid domains $VD_i$
from the starting domains $D_i$. The user-configurator interaction
can be modelled as a sequence of these executions over restricted
BDDs $B^\rho$, where the valid domains are $D_i^\rho$ and the
starting domains are $D_i^{\rho_{old}}$.

% For every variable $x_i \in X$, old domains
%are denoted as $D_i', i=0,\ldots,n-1$. The $CVD$ functionality is
%to extract values for new valid domains $D_i, i=0,\ldots,n-1$ from
%the newly restricted BDD $B(V,E,X_b,var)$.

The $CVD$ functionality is delivered by executing two algorithms
presented in Fig. \ref{CVD-Skipped} and Fig. \ref{CVD-classic}.
The first algorithm is based on the key idea that if there is an
edge $e=(u_1,u_2)$ crossing over $V_j$, i.e.
$var_1(u_1)<j<var_1(u_2)$ then we can include all the values from
$D_j$ into a valid domain $VD_j \leftarrow D_j$.

 We refer to $e$ as a \emph{long edge} of length $var_1(u_2) - var_1(u_1)$. Note that
 it skips $var(u_2) - var(u_1)$ Boolean variables,
 and therefore compactly represents the part of a
 solution space of size $2^{var(u_2) - var(u_1)}$.

\begin{figure}[htbp!]
\ttfamily

\ \     $CVD-Skipped(B)$

\ \   1: for each $i=0$ to $n-1$

\ \   2:  \ \ $L[i] \leftarrow i+1$

 \ \  3: $T \leftarrow TopologicalSort(B)$

\ \   4: for each $k=0$ to $|T|-1$

\ \   5:  \ \ $u_1 \leftarrow T[k]$, $i_1 \leftarrow var_1(u_1)$

\ \   6:  \ \ for each $u_2 \in Adjacent[u_1]$

\ \   7:  \ \ \ \ $L[i_1] \leftarrow max\{L[i_1],var_1(u_2)\}$

\ \   8: $S \leftarrow \{ \}$, $s \leftarrow 0$

\ \   9: for $i =0$ to $n-2$

\ \   10:  \ \  if $i+1 < L[s]$

\ \   11:  \ \ \ \ $L[s] \leftarrow max\{L[s],L[i+1]\}$

\ \   12:  \ \ else

\ \   13:  \ \ \ \ if $s+1 < L[s] \ \ S \leftarrow S \cup \{s\}$

\ \   14:  \ \ \ \  $s \leftarrow i+1$

\ \   15:  for each $j \in S$

\ \   16:  \ \  for $i =j$ to $L[j]$

\ \   17:  \ \  $VD_i \leftarrow D_i$

 \rmfamily
 \caption{\label{CVD-Skipped}  In lines 1-7 the $L[i]$ array is created to record longest edge $e=(u_1,u_2)$ originating
 from the $V_i$ layer, i.e. $L[i]=max\{var_1(u') \mid \exists (u,u') \in E.
 var_1(u)=i\}$. The execution time is dominated by
 $TopologicalSort(B)$ which can be implemented as depth first
 search in $O(|E|+|V|) = O(|E|)$ time.  In lines 8-14, the overlapping long segments have been merged in
 $O(n)$ steps. Finally, in lines 15-17 the valid domains have been
 copied in $O(n)$ steps. Hence, the total running time is $O(|E| +
 n)$.}
% Hence, if for each edge $e$ we define the set of skipped variables
% $X(e) = \{var_1[v_1]+1, \ldots, var_1[v_2]-1\}$, then the worst-case execution time is $O(\sum_{e \in E}|X(e)|) = O(|E| \cdot
% n)$}.
\end{figure}

%%%\begin{figure}[htbp!]
%%%\ttfamily
%%%
%%%\ \     $CVD-Skipped(B)$
%%%
%%%\ \   1: $T \leftarrow TopologicalSort(B)$
%%%
%%%\ \   2: for each $i=0$ to $|T|-1$
%%%
%%%\ \   3:  \  $v_1 \leftarrow T[i]$
%%%
%%%\ \   4:  \  for each $v_2 \in Adjacent[v_1]$
%%%
%%%\ \   5:  \   \ for each $k=var_1[v_1]+1$ to $var_1[v_2]-1$
%%%
%%%\ \   6:  \   \ \ \ $D_k' \leftarrow D_k$
%%%
%%%\rmfamily
%%% \caption{\label{CVD-Skipped} If there is an edge $e(v_1,v_2)$ crossing over $V_k$, i.e. $var_1(v_1)<k<var_1(v_2)$ then
%%% all the values from $D_k$ are valid: $D_k' \leftarrow D_k$. The lines 5 and 6
%%% are executed at most once for every edge in the graph. Hence, if for each edge $e$ we define the set of skipped variables
%%% $X(e) = \{var_1[v_1]+1, \ldots, var_1[v_2]-1\}$, then the worst-case execution time is $O(\sum_{e \in E}|X(e)|) = O(|E| \cdot
%%% n)$}.
%%%\end{figure}

\begin{figure} [htbp]
\ttfamily

\ \    $CVD(B, x_i)$

\ \   1:  \ $VD_i \leftarrow \{ \}$

\ \   2:  \ for each $j=0$ to $|D_i|-1$

\ \   3:  \ \ \ for each $k=0$ to $|In_i|-1$

\ \   4:  \ \ \  $u \leftarrow In_i[k]$

\ \   5:  \ \ \  $u' \leftarrow Traverse(u,j)$

\ \   6:  \ \ \ if $u' \neq T_0$

\ \   7:  \ \ \ \ \ $VD_i  \leftarrow VD_i \cup \{j\}$

\ \   8:  \ \ \ \ \ Return

 \rmfamily
 \caption{\label{CVD-classic} Classical CVD algorithm. $enc(j)$ denotes the binary encoding
 of number $j$ to $k_i$ values $v_0,\ldots,v_{k_i-1}$. If $Traverse(u, j)$ from Fig. \ref{CVD-Traverse}
 ends in a node different then $T_0$, then $j \in VD_i$.
 }
\end{figure}

For the remaining  variables $x_i$, whose valid domain was not
copied by $CVD-Skipped$, we execute $CVD(B,x_i)$  from Fig.
\ref{CVD-classic}. There, for each value $j$ in a domain $D_i'$ we
check whether it can be part of the domain $D_i$. The key idea is
that if $j \in D_i$ then there must be $u \in V_i$ such that
traversing the BDD from $u$ with binary encoding of $j$ will lead
to a node other than $T_0$, because then there is at least one
satisfying path to $T_1$ allowing $x_i = j$.

\begin{figure} [htbp]
\ttfamily

\ \    $Traverse(u, j)$

\ \   1:  \  $i \leftarrow var_1(u)$

\ \   2:  \  $v_0, \ldots, v_{k_i-1} \leftarrow \ enc(j)$

\ \   3:  \  $s \leftarrow var_2(u)$

\ \   4:  \  if $Marked[u]=j$ return $T_0$

\ \   5:  \  $Marked[u] \leftarrow j$

\ \   6:  \  while $s \leq k_i-1$

\ \   7:  \ \ \  if $var_1(u)>i$ return $u$

\ \   8:  \ \ \  if $v_s=0$ $u \leftarrow low(u)$

%\ \   9:  \ \ \ \ \ $u \leftarrow low(u)$

\ \   10:   \ \ else $u \leftarrow high(u)$

%\ \   11:   \ \ \ \ $u \leftarrow high(u)$

\ \   12:  \ \   if $Marked[u]=j$ return $T_0$

\ \   13:   \ \  $Marked[u] \leftarrow j$

\ \   14:  \ \   $s \leftarrow var_2(u)$

 \rmfamily
 \caption{\label{CVD-Traverse} For fixed $u \in V, i=var_1(u)$, $Traverse(u, j)$ iterates through $V_i$ and returns the node
 in which the traversal ends up.
 }
\end{figure}

%If a node $u_1$ is not on a traversal path encoding $j$ from some
%node $u$, then it is not encoding $j$ in traversal path from any
%other node $u'$.
%
When traversing with $Traverse(u, j)$  we mark the already
traversed nodes $u_t$ with $j$, $Marked[u_t] \leftarrow j$ and
prevent processing them again in the future $j$-traversals
$Traverse(u', j)$. Namely, if $Traverse(u, j)$ reached $T_0$ node
through $u_t$, then any other traversal  $Traverse(u', j)$
reaching $u_t$ must as well end up in $T_0$. Therefore, for every
value $j \in D_i$, every node $u \in V_i$ is traversed at most
once, leading to worst case running time complexity of $O(|V_i|
\cdot |D_i|)$. Hence, the total running time for all variables is
$O(\sum_{i=0}^{n-1}|V_i| \cdot |D_i|)$.

The total worst-case running time for the two  $CVD$ algorithms is
therefore $O(\sum_{i=0}^{n-1}|V_i| \cdot |D_i| + |E| +
 n) = O(\sum_{i=0}^{n-1}|V_i| \cdot |D_i| + n)$. %In all practical instances,
% the number of finite domain variables $n$ is usually much smaller then
% $V$ and the worst-case complexity is bounded by  $O(\sum_{i=0}^{i=n-1}|V_i| \cdot
% |D_i|)$.

\bibliography{all}
\bibliographystyle{splncs}

\end{document}